\pdfoutput=1
\documentclass[letterpaper, 10 pt, conference]{ieeeconf}  

\IEEEoverridecommandlockouts                              

\usepackage{graphicx}
\usepackage{epsfig} 
\usepackage{mathptmx} 

\usepackage{times} 
\usepackage{amsmath} 
\usepackage{amssymb}  
\usepackage{mathtools}
\usepackage{xcolor}
\usepackage{eucal}
\usepackage{indentfirst}
\usepackage{algorithm,algorithmic}
\usepackage{physics}
\usepackage{multirow}
\usepackage{multicol}

\usepackage{subcaption}
\usepackage{caption}
\usepackage{floatflt}

\usepackage{bm}
\usepackage{booktabs}
\usepackage{wrapfig}

\usepackage{soul}
\usepackage[normalem]{ulem}

\usepackage{pifont}
\newcommand{\cmark}{\ding{51}}%
\newcommand{\xmark}{\ding{55}}%
\newcommand{\eg}{\textit{e}.\textit{g}.~}
\usepackage{hyperref}

\usepackage{gensymb}

\begin{document}
\title{\LARGE \bf
Self-Supervised Object Goal Navigation with In-Situ Finetuning}

\author{So Yeon Min$^{1,2}$, Yao-Hung Hubert Tsai$^{1}$, Wei Ding$^{1}$,\\
Ali Farhadi$^{1}$, Ruslan Salakhutdinov$^{2}$, Yonatan Bisk$^{2}$, Jian Zhang$^{1}$%
\thanks{The authors are with $^1$ Apple and $^2$ Carnegie Mellon University, USA. Correspondence: {\tt\small soyeonm@andrew.cmu.edu}
} 
}

\maketitle

\begin{abstract}

A household robot should be able to navigate to target objects without requiring users to first annotate everything in their home.
Most current approaches to object navigation do not test on real robots and rely solely on reconstructed scans of houses and their expensively labeled semantic 3D meshes. 
In this work, our goal is to build an agent that builds self-supervised models of the world via exploration, the same as a child might - thus we (1) eschew the expense of labeled 3D mesh and (2) enable self-supervised \textit{in-situ} finetuning in the real world.
We identify a strong source of self-supervision (\textit{Location Consistency - LocCon}) that can train all components of an ObjectNav agent, using unannotated simulated houses.
Our key insight is that embodied agents can leverage location consistency as a self-supervision signal -- collecting images from different views/angles and applying contrastive learning. We show that our agent can perform competitively in the real world and simulation.
Our results also indicate that supervised training with 3D mesh annotations causes models to learn simulation artifacts, which are not transferrable to the real world. In contrast, our \textit{LocCon} shows the most robust transfer in the real world among the set of models we compare to, and that the real-world performance of all models can be further improved with self-supervised \textit{LocCon} in-situ training.

\end{abstract}

\section{Introduction}
\label{sec:intro}

In practical deployment scenarios where a robot is delivered to a customer's house, 
it should quickly generalize and \textit{adapt} to new objects and layouts.
For the task of object goal navigation (ObjectNav), this means that the robot should adapt both 1) its visual perception (\textit{which pixel is which object?}) to the objects and 2) the navigation policy (\textit{where to look next?}) to the house layout in the deployment setting. Because this should happen automatically, without the user annotating data, we propose \textit{in-situ} learning (i.e. learning in the deployed \textit{real-world} setting) of ObjectNav agents. Our proposed \textit{in-situ} adaptation is not applicable to existing approaches \cite{ogn, poni} as they depend on expensive labeled semantic 3D meshes of scanned and reconstructed houses to train their agents and perception stacks.

\begin{figure}[!t]
    \centering
    \includegraphics[width=0.47 \textwidth]{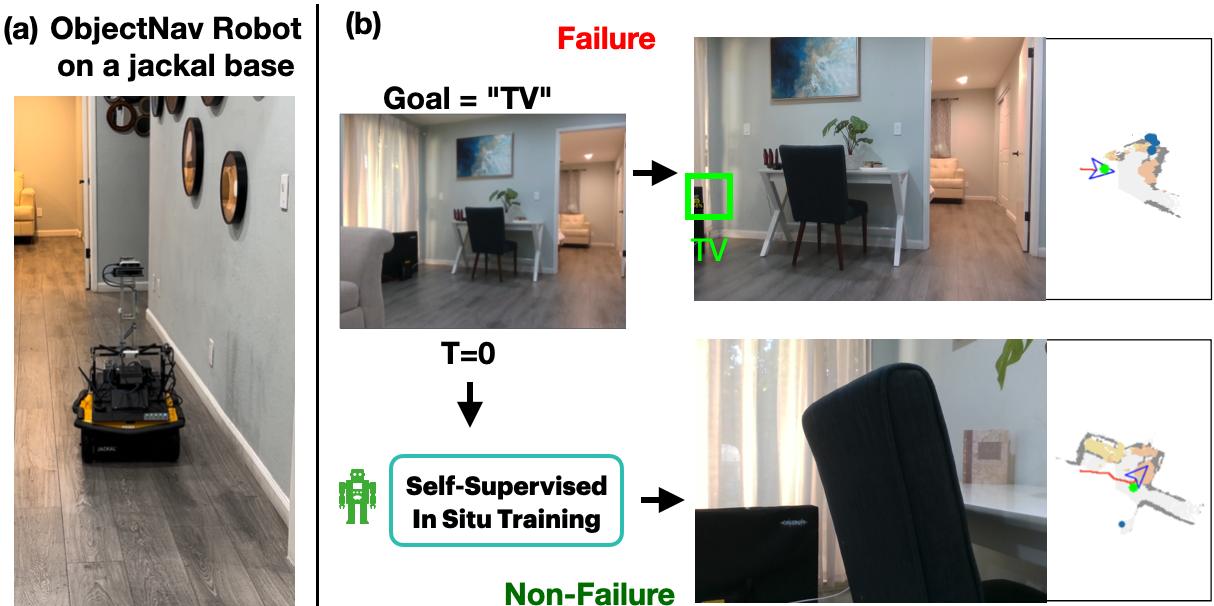}
    \caption{\textbf{Real World ObjectNav and \textit{in-Situ} Training} (a) Picture of our robot in one of the airbnbs rented for ObjectNav experiments. (b) Example ObjectNav (real-world) failure mode and its remedy with \textit{in-situ} self-supervised (location consistency) training. The agent starts at the living room and wrongly detects a small patch of a black object as ``TV.'' The segmentation model avoids this distractor after \textit{in-situ} location consistency training.}
    \label{fig:insitu}
\end{figure}

For this purpose, we introduce a novel and strong source of self-supervision (\textit{Location Consistency - LocCon}), and propose a method to train all components of an ObjectNav agent \textit{without} labeled 3D meshes.
At a high level, our approach is composed of two stages: Stage I (visual perception) utilizes the embodiment of an agent to train a semantic segmentation model, and then Stage II (Nav Policy) uses the trained semantic segmentation model to self-label the 3D mesh, which then can be used to train an ObjectNav policy. In particular, in Stage I, we use location consistency as a self-supervision signal: an agent collects images from different views/angles for a given location. We scale the location-consistency collection to 100 unlabeled houses and apply contrastive training to fine-tune the backbone of a pretrained semantic segmentation model \cite{deeplab} (\textit{LocCon} training). In Stage II, we train a navigation policy inspired by PONI \cite{poni}, which learns potential functions over map frontiers (analytic functions of unexplored area and geodesic distances to goal locations), with self-labeled semantic maps.

Our analysis finds that (1) our fully self-supervised ObjectNav agent can perform competitively in the real world, (2) \textit{in-situ} training with \textit{LocCon} further improves visual perception performance at deployment, and (3) despite the presence of noisy artifacts in simulation, our self-supervised training can learn useful semantics for sim2real transfer.  
We present a comprehensive set of experiments in real houses and simulation (Gibson ObjectNav datasets \cite{ogn, GibsonEnv, zer}). 

First, we perform a sim2real experiment that shows our agent trained in simulation can be directly run in the real world for object goal navigation, demonstrating the robustness of our approach. We find that the largest error mode of real-world ObjectNav is visual perception errors, which often \textit{nullifies} the need to adopt a better ObjectNav policy. This analysis, together with evidence from previous work \cite{ogn} that the Nav Policy, captures abstract knowledge of house layouts which transfers to the real world, shows that \textit{in-situ} training is most needed for updating visual perception. 
Second, we perform self-supervised location-consistency training directly in the {\em real houses} -- a step towards agents learning on their own. In particular, with location consistency data collected from a {\em real robot} from 4 AirBnB houses, we show that segmentation models can be improved with our training scheme. Finally, within simulation, we compare our self-supervision pipeline against training with annotations (\textit{FullSup} agent). We find that \textit{FullSup}'s segmentation model learns the visual noise from simulation - which is irrelevant and harmful for real-world transfer. This causes \textit{FullSup} to perform \textit{worse} and makes it a harmful initialization in the real world. 

Our first-of-a-kind full stack system for self-supervised ObjectNav enables us to investigate where simulation is most helpful or harmful in training real world ObjectNav agents. The primary approach in the literature has been to focus on collecting scans of houses for reconstruction and training in simulation; however, our results indicate that simulated data often introduces both reconstruction and annotation noise, thereby inhibiting real world transfer. Where supervised learning incentives models to overfit to such artifacts, we find that our self-supervised training is robust to reconstruction errors and help real world transfer. This disparity begs the question - \textit{What data collection and annotation efforts are most effective in the real world and simulation?} We argue, that while simulation holds promise, in the absence of high-quality annotations or renderings, researchers should adopt our self-supervised approach which can be applied directly in the real world for \textit{in-situ} training with raw, unannotated, images that can be easily collected.

\begin{figure*}[!t]
    \vspace{1em}
    \centering
    \includegraphics[width=1.0\textwidth]{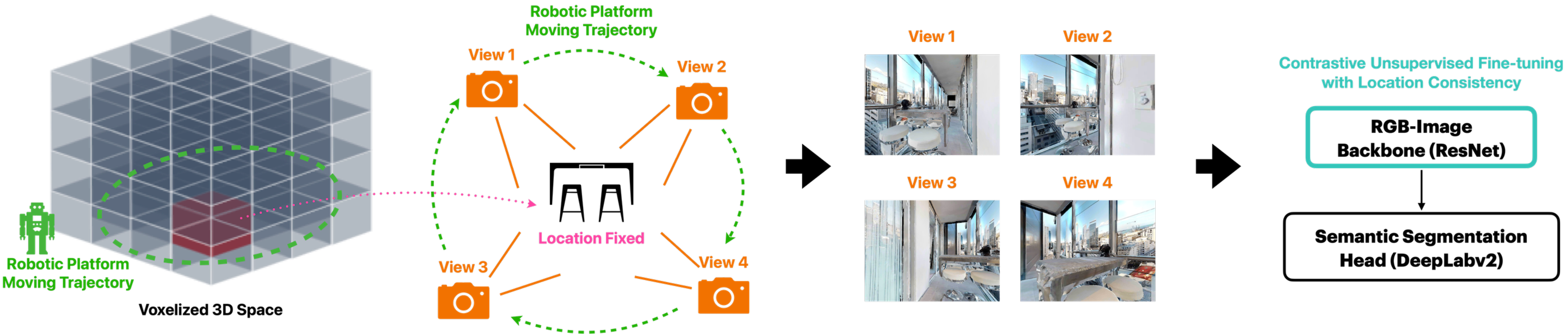}
    \vspace{-1.5em}
    \caption{\textbf{Fine-tuning visual perception using self-supervision:}
    First, we produce a 3D voxel grid of the scene from depth sensors of a robot. We randomly select a location and put 50cm$^3$ cube, and make the robot view it from 8 different angles and 40 different poses. If the ray reflects back from this cube to the robot's camera, we save the corresponding egocentric RGB frame. These collected images are labeled with ``location consistency'' pseudo-labels in a self-supervised manner. We also show examples of collected ``location consistent'' images. The ResNet backbone of pre-trained semantic segmentation models can be fine-tuned with our data using contrastive loss.}
    \label{fig:selfsup_visual}
\end{figure*}

\section{Related Work}
\label{sec:rel}

We enumerate related work for visual perception learning and policy learning. We will transition the discussions from supervised learning, which requires a lot of annotated labels, to self-supervised learning, which avoids labels. 

{\bf Visual Representation Learning.} In ObjectNav, ground truth 3D semantic labels are often given in the dataset and simulators, such as Matterport3D~\cite{chang2017matterport3d}, Replica~\cite{straub2019replica}, Habitat~\cite{savva2019habitat}, Gibson~\cite{GibsonEnv}, iGibson~\cite{li2021igibson}, and Habitat-Matterport3D~\cite{ramakrishnan2021hm3d}. Existing methods often leverage the ground truth semantic 3D labels as strong supervision signals for ObjectNav~\cite{ogn,poni}. Nonetheless, because of lacking real-world semantic labels, it is unclear if the methods will generalize well from the simulator to the real world.

\footnotetext{While \cite{ogn}, a closely related work of \cite{SEAL}, shows real world transfer of their semantic mapper and semantic nav policy, \cite{SEAL} itself does not show real world transfer results of its self-supervised components.}

To avoid the need for ground truth 3D semantic labels, previous work has proposed workarounds: Mimic behaviors from a large collection of human demonstrations~\cite{habitat-web},
gather large-scale image collections from 3D mesh and apply SSL techniques for static images~\cite{ovrl},  
or leverage the embodiment of a moving agent for self-supervised learning by taking the action of the agent and the interaction with the environment into account when building their representations~\cite{crl, SEAL, evr}.\footnote{Most self-supervised learning approaches leverage only static images without considering the embodiment of an agent. SimCLR\cite{simclr}, MoCo \cite{moco}, DINO \cite{DINO}, and InfoMax\cite{infomax} build consistent representations for augmented variants of an same image and maximize their difference to other images.
} For example, in SEAL \cite{SEAL} an agent roams in an environment, records predictions from an initial semantic segmentation model, propagates the predictions from high-confidence to low-confidence regions, and finally uses the predictions to fine-tune the semantic segmentation model. The resulting model is then used for navigation. In principle, SEAL and LocCon could be used together; however, LocCon is more general as, unlike SEAL, it does \textit{\textbf{not}} assume a pre-defined category of objects for training. In CRL~\cite{crl}, the agent learns a policy to search for images that are deemed helpful for contrastive learning. 
In both SEAL and CRL, self-supervisedly collecting data requires a new training/calibration of policies - which can be a hindrance to real-world transfer; our method, on the other hand, uses algorithmic data collection upon a 3D map, which is simpler and does not require learning. The simplicity and independence of our method allows for easy transfer of our self-supervised learning scheme in the real world (\S\ref{sec:real_insitu}).

\begin{table}[t!]
\renewcommand{\arraystretch}{1.2}
\label{tab:comparisons}
\centering
\vspace{1ex}
\begin{footnotesize}
\begin{tabular}{@{}l@{\hspace{5pt}}c@{\hspace{5pt}}c@{\hspace{5pt}}c@{\hspace{5pt}}c@{}}
\toprule
       & SSL Visual & SSL Nav & Real Robot & Real \\
Method & Perception & Policy  & Transfer   & World SSL\\
\midrule
SEAL~\cite{SEAL} & \textcolor{green}{\cmark} & \textcolor{red}{\xmark} & \textcolor{red}{\xmark}$^*$ & \textcolor{red}{\xmark} \\
OVRL~\cite{ovrl} & \textcolor{green}{\cmark} & \textcolor{red}{\xmark} &\textcolor{red}{\xmark} & \textcolor{red}{\xmark}\\
ZER ~\cite{zer} & \textcolor{green}{\cmark} & \textcolor{green}{\cmark} &  \textcolor{red}{\xmark} &  \textcolor{red}{\xmark} \\
CoW, ZSON~\cite{cow, zson} & \textcolor{green}{\cmark} & \textcolor{green}{\cmark} &  \textcolor{red}{\xmark} &  \textcolor{red}{\xmark} \\
\midrule
Ours & \textcolor{green}{\cmark} & \textcolor{green}{\cmark}&\textcolor{green}{\cmark} & \textcolor{green}{\cmark} \\
\bottomrule
\end{tabular}
\end{footnotesize}
\caption{\textbf{Comparison of self-supervised ObjectNav methods.} Our work (1) employs end-to-end self-supervised training both for the visual perception and navigation policy, and (2) demonstrates the sim2real transfer of both components to the real world.\protect\footnotemark
}
\vspace{-2em}
\end{table}

{\bf Policy Learning.} In ObjectNav, policies fall into two categories: end-to-end and modular approaches. End-to-end methods \cite{etoe1, etoe2, etoe3, etoe4, procthor, embclip} directly predict low-level actions (e.g. ``move forward'', ``turn left'') from input RGB-D images and the camera pose. Modular methods \cite{ogn,poni,zson,ovrl, cmp} consist of a pipeline of components - semantic mapping (persistent semantic and spatial memory), a high-level semantic policy (decides the general direction to move towards the predicted goal location), and low-level navigation modules (a low-level planner for navigating to the high-level goal chosen by the semantic policy). Learning is typically constrained to 1) the semantic (instance) segmentation model inside the semantic mapper and 2) the high-level semantic policy; our method is modular.

Existing methods for semantic policy training often use the object goal's ground truth location, either to define reward functions for reinforcement learning \cite{ogn, ovrl} or potential functions for supervised learning \cite{poni} - 
these are expensive requirements that 
make these approaches difficult to train in the real world. To avoid the need for ground truth object goal locations, ZER, ZSON, and CoW~\cite{zer, zson, cow} perform \textit{image} goal navigation (navigating to a target image rather than a category). These images are converted to goal embeddings via CLIP~\cite{radford2021learning} or ResNet\cite{resnet}-based encoders, allowing for object categories to be mapped to scenes in service of performing ObjectNav.
In lieu of web-scale pretraining, given an object goal  
our method learns to find the object in our self-supervised semantic 3D map without the use of any ground truth location information. 

While not related to self-supervision, \cite{procthor} uses data at scale to improve performances in ObjectNav. More specifically, it trains EmbCLIP~\cite{embclip} on 10K diverse houses procedurally generated. While we use more homogenous scenes from Gibson~\cite{GibsonEnv} and HM3D~\cite{hm3d}, the implication of our work well agrees with \cite{procthor} in that scaling brings gains. On the other hand, \cite{gervet2022navigating}, a concurrent work, presents large-scale comparisons of ObjectNav methods when transferred to the real world. However, the focus of \cite{gervet2022navigating} and our work are different in that we focus on the need of/method for self-supervised real-world \textit{in-situ} training.

\begin{figure*}[!t]
    \centering
    \includegraphics[width=0.95\textwidth]{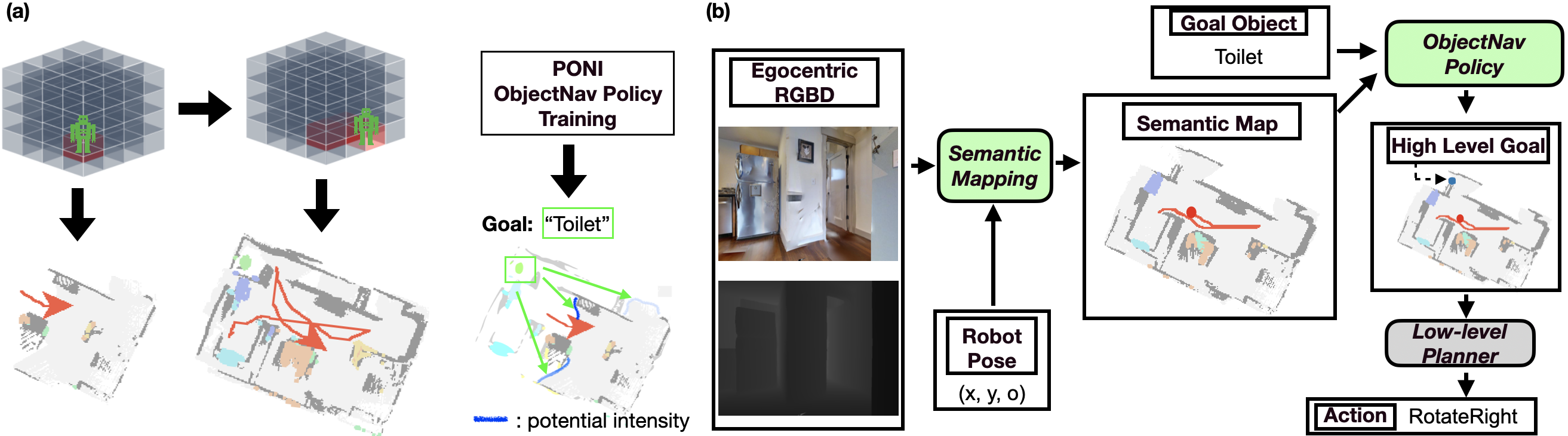}
    \caption{\textbf{ObjectNav policy training and pipeline.} (a) Self-supervised ObjectNav policy training: The agent constructs self-labeled semantic maps of the environment using the segmentation model trained in Stage I. Then, we create a partial map out of full maps, and compute potential functions upon the partial map from the unexplored areas and locations of goal objects. 
    Finally, we apply PONI policy training with the self-labeled potential functions. (b)  ObjectNav agent pipeline: Following previous work~\cite{ogn,poni, min2021film}, we 
    equip our ObjectNav agent with the semantic mapper and the Object Nav policy. The learned components are colored in green.}
    \label{fig:selfsup_policy}
\end{figure*}

\section{Task Explanation}
\label{sec:task_explanation}

An ObjectNav \cite{anderson2018evaluation, savva2017minos} agent aims to navigate to an instance of the given object category such as ``table'' or ``sofa'' as efficiently as possible, with its perception of the world and semantic priors. The agent is placed at a random location in the environment and receives the goal object category (e.g. ``table''). With each action ($a_t$) the agent takes, it records visual observations ($V_t$) and pose information ($x_t$). 

In this work, we perform ObjectNav both on real robots and within simulation.  This leads to several key changes in the action space, and sensor noise in both visual observations and pose information. In both conditions, visual observations will consist of first-person RGB-D images. 

\paragraph{Simulation}
Within simulation, we have an idealized action space $A$ that contains four actions: \texttt{forward} (0.25 meters), \texttt{left} (30\degree), \texttt{right} (30\degree), and \texttt{stop}. The agent takes the \texttt{stop} action when it believes it has reached the goal. The episode is considered successful, if the agent's final location is within a distance $1m$ of the goal.  Additionally, we set a maximum number of timesteps (500).

\paragraph{Robot Platform} For the real-world Object Nav, we rented three AirBnB's in North America, using the same set of object categories used in simulation. Using the fast-marching method, the robot plans collision-free paths connecting its position to the desired goal location; then, using the \texttt{ROS move\_base} package, it regulates velocities and yaw rates to follow the planned paths. Since 0.25 meters and 30 degrees are too small for velocity/yaw rate control, we directly use  $(x, y, o)$ (where $x, y$ are the “forward” and “left” positions from the starting pose in meters and $o$ is the orientation in radian) as the action space. After each time step, we pass the next desired global pose to \texttt{move\_base}.
To account for velocity control and the body radius, the task is considered successful if the robot stops within 1.5 meters of the desired goal. Since the action space is no longer discrete, we set a maximum duration of time (7 minutes) instead of timesteps. Furthermore, to account for localization errors, we allow up to two trials per task. Additional robot hardware specifications are in \S\ref{subsec:real_world_nav}.

\section{Methods}
\label{sec:Methods}

Our agent is composed of a semantic mapper and a navigation policy (Fig.~\ref{fig:selfsup_policy} (b)), following previous work \cite{ogn, poni}. Mapping converts egocentric RGB-D and robot pose information to a 5cm$^3$ voxel representation of the scene, with labels for occupancy, explored areas, and semantic categories for each voxel \cite{ogn}. Core to mapping, is learning semantic (instance) segmentation, which enables object category labeling. 
The navigation policy (Fig.~\ref{fig:selfsup_policy} (a)) is responsible for spatial understanding and deciding where to look/move next.

\subsection{Location Consistency for Visual Perception}
\label{subsec:Methods_loccon}

The self-supervised finetuning of the semantic segmentation model is composed of two steps (Fig.~\ref{fig:selfsup_visual}). First, we algorithmically program a robot to navigate around objects to link images with location consistency. Second, we train the model in an off-line manner, using a contrastive loss.  

\paragraph{Location consistency data collection}
In a given scene, we initiate an agent and create a 3D occupancy map ($O$) of the environment with frontier-based exploration \cite{fbe}; the robot continues to move to unexplored areas of the map until there is no unexplored area left. 
Once complete, we use $O$ to sample the (next) location that the agent will observe from multiple views (Fig.~\ref{fig:selfsup_visual}). First, we sample a point $(x, y)$ among the occupied regions ($O$ summed across height). Second, from $O[x,y]$, we sample $z$ (height) among occupied voxels. Third, we place a 50 cm$^3$ cube $C$ centered around $(x,y,z)$ and have the agent move around this cube at increment angles of 45$^\mathrm{o}$ (i.e. 0, 45, 90, ...) and distances of $0.3, 0.6, 0.9, 1.2$ meters from $(x,y,z)$. ``Intermediate stop'' poses (where the agent stops temporarily to stare at $C$) are placed in navigable regions of $O$. When the agent pauses at an ``intermediate stop'', it ray casts and if it hits $C$ 
the current egocentric RGB view $R$ is recorded as an image of $C$. When the agent has visited all ``intermediate stops'' for the current $C$, it saves all the views as a ``location-consistent'' datum. The agent then repeats the entire process by sampling and collecting data from a new location $(x,y)$ from $O$ until 70$\%$ of the map is covered or the agent has moved 7,000 steps. 

We conduct the same data collection process in multiple real houses to obtain a collection of images. We only consider images from the same cube $C$ as location-consistent.  In the physical environment, there are additional complications as agents cannot achieve perfect angles and distances, but the procedure still succeeds and is not adversely affected by these minor perceptual changes. We enumerate factors that could influence the data collection result, such as camera extrinsics, architectural choices, and robot morphology; we hope future work explores these questions.

\paragraph{Contrastive training (LocCon)}

As with previous work, we initialize our backbone with a pre-trained semantic segmentation model, but training then commences using the aforementioned location consistency data and contrastive training.  The location serves as the label tieing together images from disparate views (Fig.~\ref{fig:selfsup_visual}). 

We denote a 
model under training as $M$ (backbone and segmentation head), and its backbone as $M_b$. We denote ${ I}$ as an image, $\{{ I}^k_{{\rm pos}, i}\}_{i=1}^n$ as $n$ images collected from the same $k$-th cube $C_k$ or augmented variants of these images, $\{{ I}^k_{{\rm neg}, j}\}_{j=1}^m$ as $m$ images not collected from $C_k$, ${\rm sim}(u, v) = M_b(u)^\top M_b(v)/ \|M_b(u) \| \|M_b(v) \|$ as the cosine similarity between two embedding vectors $M_b(u)$ and $M_b(v)$, and $\tau$ as a temperature hyper-parameter. We consider the popular contrastive loss as in prior work~\cite{moco,simclr}:
\begin{equation}
\ell_{B} = \sum_{k=1}^K\sum_{i=1}^n\,-{\rm log}\,\frac{e^{{\rm sim}({I}, {I}_{{\rm pos}, i})/ \tau }}{e^{{\rm sim}({I}, {I}_{{\rm pos}, i})/ \tau }+\sum_{j=1}^m e^{{\rm sim}({I}, {I}_{{\rm neg}, j})/ \tau }}.
\label{eq:eq1}
\end{equation}

To synchronize the segmentation head with the pretrained backbone, we experimented with the addition of a regularizer, but we found no effect on downstream performance. 
For each minibatch, the loss is therefore simply $\ell_{B}$. 
We select $64$ different locations in an environment, and for each location we sample four images from different views and apply image augmentations to obtain two augmented variants of each of the four images. This construction, with a mini-batch of 512 images, results in each image having seven positively-paired images ($n=7$) and $504$ negatively-paired image ($m=504$). We are interested in the semantic segmentation of 15 semantic categories Table~\ref{tab:sim-seg-obj}), following previous work \cite{ogn, poni}. Since the pre-trained model comes with a segmentation head for more than 15 classes, we treat the rest as ``background.'' Adam optimizer with weight decay $1e-4$ and the learning rate $\in \{1e-6, 1e-5, 1e-4\}$ that outputs the best validation accuracy ($1e-5$) was used.

\subsection{A navigation policy from self-labeled scenes}
\label{subsec:Methods_nav_policy}

The ObjectNav policy determines the next intermediate stop to find the goal, based on the partial observations of the agent (Fig.~\ref{fig:selfsup_policy} (a)). Unlike in prior work
\cite{poni}, 
we demonstrate how training can be done without a labeled navigation mesh. 

We follow a process similar to the pre-mapping of \S\ref{subsec:Methods_loccon}, but now also record semantic information. We equip an agent with the semantic segmentation model trained in \S\ref{subsec:Methods_loccon}, and enforce frontier-based exploration; this creates a semantic map of the scenes, self-labeled by the visual perception model. Then, we 
learn the ``potential functions'' over the frontiers of partial maps; a high potential function at a point in the map indicates that the point is worth visiting. The total potential function is a sum of the ``area potential function'' (high intensity means that the point is where the agent should ``explore'' to gain a more complete map) and the ``object potential function'' (high intensity means that the point is likely to be close to the object being looked for). The area and object potentials can be directly calculated from the geodesic distances obtained from the \textit{``full map''}, and the policy network learns the potentials given a \textit{``partial map,''} similar to one that the agent will create during a task.

\begin{figure}[!t]
    \centering
    \includegraphics[width=0.5\textwidth]{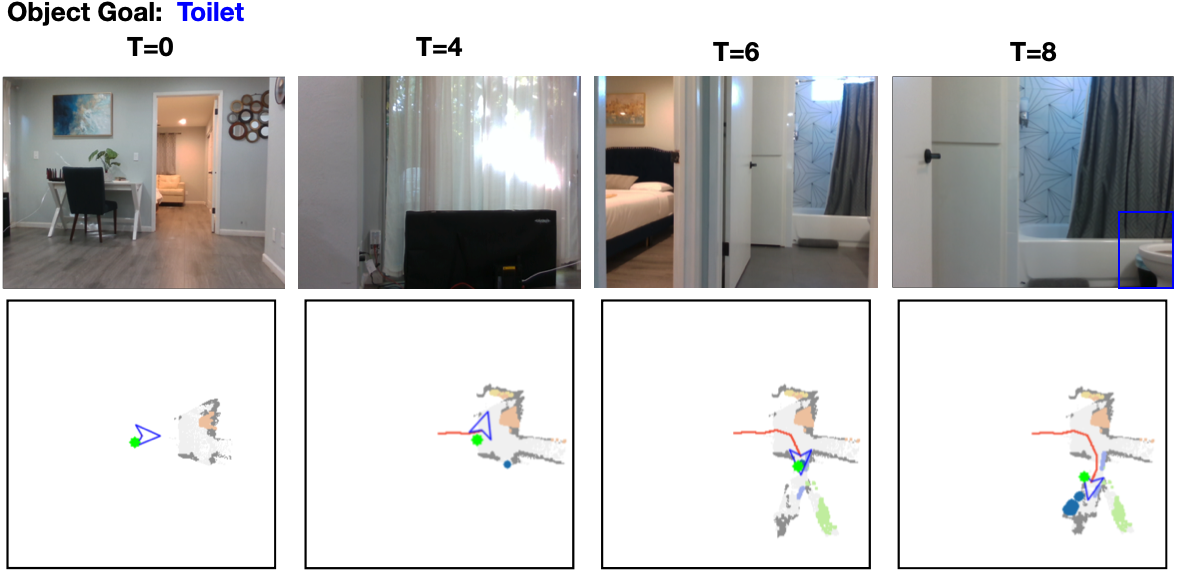}
    \caption{\textbf{Example ObjectNav task in a real house.} The goal object is toilet. Visual perception, semantic mapping, and the ObjecNav policy were all accurate enough to lead to a task success.}
    \label{fig:real_robot_experiment}
\end{figure}

\section{Real-World Experiments}
\label{sec:Nav_exp}

In \S\ref{subsec:real_world_nav}, we first show that our agent, trained with self-supervision only, can perform ObjectNav competitively in the real world and in simulation; in \S\ref{sec:real_insitu}, we show that self-supervised training of visual perception results in good real world transfer, and real-world \textit{in-situ} training can further remedy ObjectNav error modes.  

The structure of our ObjectNav agent is shown in Fig.~\ref{fig:selfsup_policy} (b). The agent relies on our self-supervised segmentation model (\textit{Self-Seg}) and ObjectNav Policy (\textit{Self-Pol}). \textit{Self-Seg} is trained by taking an ``off-the-shelf" model (aka deeplabv2\cite{deeplab} pre-trained with COCO-things 164k\cite{coco}) and applying LocCon (\S\ref{subsec:Methods_loccon}) with data sampled from 100 houses, randomly selected from a pool of 25 houses from the training split of Gibson-mini and 720 houses from HM3D \cite{hm3d}. We again highlight that \textit{no labeled 3d mesh} was used in this training, since they are expensive and mostly unavailable for HM3D.
\textit{Self-Pol} is trained using the method of \S\ref{subsec:Methods_nav_policy}, from semantic maps created and self-labeled (with \textit{Self-Seg}) from frontier-based exploration in the 25 Gibson houses. 

We define notations used in the remaining sections. We denote the ``off-the-shelf" segmentation model (aka deeplabv2 \cite{deeplab} pre-trained with COCO-things 164k \cite{coco}) as \textit{O-t-S.} We replace \textit{Self-Seg} and \textit{Self-Pol} with their fully supervised (using all the available simulator annotations) counterparts; we call this agent ``\textit{FullSup} agent'' (and its components \textit{Full-Seg} and \textit{Full-Pol})  and our self-supervised agent ``\textit{SelfSup} agent.'' \textit{Full-Seg}, initialized with \textit{O-t-S.}, is trained with fully supervised training from 75K randomly sampled, labeled images from the 25 Gibson houses, and \textit{Full-Pol} is trained with PONI using the ground-truth labeled 3d mesh of the 25 houses. We follow the same protocols of \cite{deeplab} and \cite{poni} (e.g. the pool of hyperparameters to optimize within, output dimensions) to train  \textit{Full-Seg} and \textit{Full-Pol}, respectively.

\subsection{Real World ObjectNav}
\label{subsec:real_world_nav}

To show the transfer of our ObjectNav agent (\S\ref{subsec:Methods_nav_policy}, \S\ref{sub_sec:sim_nav}) to real houses, 
we built a robot from a Jackal base that has similar specifications to the simulation settings (\S\ref{sec:task_explanation}). Perception is performed with an Intel Realsense D435 RGB-D camera mounted at 0.88 meters from the floor and all computation was run locally in real-time on an Nvidia Orin (Fig.~\ref{fig:real_robot_experiment} (a)). 
\setlength{\columnsep}{5pt}%
\begin{wrapfigure}[10]{r}{0.35\linewidth}
\vspace{-1em}
\centering
\footnotesize
\fontsize{8}{8}\selectfont
\setlength\tabcolsep{4pt} 
\begin{tabular}{@{}lc@{}}
\textbf{Category}
   &\textbf{Success} \\
\toprule
\;Total 
                & \phantom{0}66.6\%      \\    %
\midrule
\;Chair 
                & \phantom{0}33.3\%    \\    %
\;Couch 
                & 100.0\%    \\    %
\;Potted Plant 
                & \phantom{0}66.6\%  \\    %
\;Bed 
                & 100.0\%  \\    %
\;Toilet 
                & 100.0\%     \\    %
\;TV 
                & \phantom{0}33.3\%  \\    %
               
\bottomrule
\end{tabular}
\captionof{table}{\textbf{ObjectNav Results} Success rate of 18 tasks in 3 AirBnBs.}
\label{tab:realworld_objectnav}
\end{wrapfigure}
For localization, we pre-map the environment with the \texttt{ROS RTAB-map} package when first entering a new house. When executing a policy, the robot receives its pose from the ROS SLAM package and uses the existing map only for its occupancy map instead of that from the semantic mapper; all other outputs of the semantic mapper, including mapping of explored areas and semantic categories, are as in simulation.

Experiments were performed in three rented AirBnBs where we set up three tasks for each of the 6 goal objects.
Because a Jackal base is larger than the agent radius in simulation, we start the agent in areas that are wider than 1.5 meters, which is usually in the living room or the kitchen. An example experiment is shown in Figure~\ref{fig:real_robot_experiment}. Across 18 tasks, we obtain success rate of $66.6\%$. We show two example runs in the supplementary video.
Out of 6 failures, 4 were due to misrecognition of goal objects (e.g. recognizing a \textit{``white couch''} as \textit{``bed''}), and 1 was due to depth/ segmentation misalignment, 1 was due to localization failure.

\subsection{Self-Supervised In-Situ Training of Visual Perception}

\label{sec:real_insitu}

In \S\ref{subsec:real_world_nav}, we saw that a major error mode in real world ObjectNav is visual perception; the robot stops at wrong goal objects or goal objects are seen but misrecognized. Since we find that visual perception generalization failures somewhat \textit{nullify} efficient exploration/ nav policy, we focus on \textit{in-situ} learning for visual perception. We show that \textit{LocCon} training of semantic segmentation can remedy these issues.

To replicate applying LocCon in a the real world, we collected location consistency images from 4 airbnbs, of which three are disjoint from the ones in \S\ref{subsec:real_world_nav}. We replicate the sampling procedure of \S\ref{subsec:Methods_loccon} as follows; we use a pre-map of the environment to sample an $(x,y)$ and then pick $z$ as in \S\ref{subsec:Methods_loccon}. We place a tripod with Intel Realsense D435 RGBD camera mounted at 0.88 meters at 5 to 10 locations around the picked $(x,y,z)$ among navigable points from which $(x,y,z)$ is visible, mimicking the ``intermediate stops'' of \S\ref{subsec:Methods_loccon}. We collected a total of 479 images and 70 $(x,y,z)$'s. For validation, we annotate five images (and exclude them from the training set) from the AirBnB that was used in \S\ref{subsec:real_world_nav}, so that all goal object classes appear at least once. We show examples of train/ val images in the supplementary video. 

In Table~\ref{tab:sem_seg_realworld}, we show the Mean IOU of \textit{O-t-S.}, \textit{Self-Seg} (Self.), and \textit{Full-Seg} (Full.), when evaluated and \textit{in-situ} trained on real world data with \textit{LocCon}. First, we find that Full. gives the worst and Self. gives the best initialization in the real world. 
This suggests that our  \textit{LocCon} training makes simulation training useful for Sim2Real transfer, despite the data challenges shown in \S\ref{sec:sim_bad}. Second, we find that our LocCon training gives performance boost for all models.
\begin{wrapfigure}[]{r}{0.5\linewidth}
\centering
\fontsize{7}{7}\selectfont
\setlength\tabcolsep{3pt} 
\begin{tabular}{@{}lccccc@{}}
\toprule
   &\textbf{O-t-S.} && \textbf{Self.} && \textbf{Full.} \\
\midrule
\;Initial
                &  61.1  &&  61.9  && 58.8   \\    
\;with In-Situ
                &  62.2   &&  62.8  &&  61.7 \\    
\midrule
\;Performance $\Delta$ 
    &  1.1 $\uparrow$   &&  0.9 $\uparrow$   &&  2.9$ \uparrow$  \\       
\bottomrule
\end{tabular}
\captionof{table}{Semantic segmentation gains from real-world \textit{in-situ LocCon} finetuning.}
\label{tab:sem_seg_realworld}
\vspace{-2em}
\end{wrapfigure}
While we do not include plots due to space limmitations, we found that as we increase the number of training images for LocCon training, the IOU increases, in both simulation and the real world. 

We retrofit \textit{Self-Seg} trained with In-Situ (Table~\ref{tab:sem_seg_realworld}) to real world ObjectNav (\S\ref{subsec:real_world_nav}) and examine if this adapted model can change any failures to non-failures (Fig.~\S\ref{fig:insitu}). More specifically, we replicate the same trajectory the agent took in each of the failed tasks, but use \textit{Self-Seg} trained with In-Situ (Table~\ref{tab:sem_seg_realworld}) in the semantic mapper. We observe that this turns one out of six failed tasks as a non-failure (Fig.~\ref{fig:insitu}).

\section{Simulation Experiments}
In \S\ref{sub_sec:sim_nav}, we present ObjectNav results in simulation. In \S\ref{subsec:errormodes}, we analyze navigation error modes, with special attention to those caused by simulation artifacts (Fig.\ref{fig:data_challenges}). In \S\ref{subsec:seg_comparison}, we provide an in-depth analysis of self-supervision vs. full supervision (with annotations) for visual perception, and conclude that the latter tends to learn artifacts from bad rendering and annotations.
In \S\ref{subsec:polcomparison}, we analyze the performance and behavior of a self-supervised nav policy. 

\label{sec:sim_bad}

\subsection{ObjectNav in simulation}
\label{sub_sec:sim_nav} 

While our focus is on the deployable real-world system, we also evaluate our agent in simulation (two validation sets of Gibson ObjectNav) for fair comparison to both existing supervised models \cite{ogn} (Gibson-1) 
and fully self-supervised approaches \cite{zer} (Gibson-2). Table \ref{tab:objectnav_results} shows ObjectNav results of our approach, compared with other fully supervised and partially (visual perception is self-supervised while policy was trained with annotated mesh)/fully self-supervised methods. Succ. denotes task success rate across all validation tasks, SPL the Succ. weighted by path length, and DTS the average distance to the 1 meter (stop threshold) radius of the true goal. Our method outperforms fully self-supervised methods on Gibson-2 and performs as competitively as methods with more supervision on Gibson-1. 
 
From Table \ref{tab:objectnav_results}, we do observe that directly training with simulation annotations gives advantages; however, we find that this in-domain advantage is due to adapting to the \textit{artifacts} and \textit{noise} of simulation, not learning the indoor semantics that is transferrable to the real world (Fig.~\ref{fig:data_challenges}).
In \S\ref{subsec:errormodes}, \S\ref{subsec:seg_comparison}, \S\ref{subsec:polcomparison}, we provide in-depth analyses of the challenges of supervised training with simulation artifacts.

\newcommand{\uarr}{$\uparrow$}
\newcommand{\darr}{$\downarrow$}
\newcommand{\code}[1]{\texttt{#1}}
\begin{table}[!t]
    \centering
    \begin{footnotesize}
    \begin{tabular}{@{}lc@{\hspace{5pt}}c@{\hspace{5pt}}cc@{\hspace{5pt}}c@{\hspace{5pt}}c@{}}
    \toprule
                                              & \multicolumn{3}{c}{Gibson - 1 \cite{anderson2018evaluation}}     & \multicolumn{3}{c }{Gibson - 2\cite{zer}}          \\ \cmidrule(l){2-4}\cmidrule(l){5-7} 
    Method                                    & Succ. \uarr &  SPL \uarr &  DTS \darr &  Succ. \uarr  & SPL \uarr \\ \midrule
    \multicolumn{7}{@{}l}{\textbf{Fully Supervised w. Annotations}} \\
    \; \code{DD-PPO}~\cite{ddppo}        &     15.0    &     10.7   &     3.24   &        N.A.     &    N.A.          \\
    \; \code{SemExp}~\cite{ogn}    &     65.5$^*$\!\!\!    &   36.5$^*$\!\!\!   &     1.45$^*$\!\!\!   &   42.8$^*$\!\!\!      &      20.8$^*$       \\
    \; \code{PONI} ~\cite{poni}                        & 66.2$^*$\!\!\!   & 36.8$^*$\!\!\!  & 1.41$^*$\!\!\!  &  34.9$^*$\!\!\!    &   20.3$^*$\!\!\!  \\     
    \midrule
    \multicolumn{7}{@{}l}{\textbf{Partially Self-Supervised}} \\
    \; \code{SemExp + SEAL}~\cite{SEAL}    &     62.7    &    33.1    &   N.A.     &        N.A.     &      N.A.      \\ \midrule
    \multicolumn{7}{@{}l}{\textbf{Fully Self-Supervised}} \\
    \; \code{ZER}~\cite{zer}    &    N.A   &    N.A  &     N.A   &     11.3      &    N.A \\
    \; \code{ZSON}~\cite{zson}     &     N.A    &     N.A   &    N.A &   31.3  &   12.0      \\
    \; \code{\textbf{Ours}}     &     \textbf{60.0}    &     31.2   &     1.89  &     \textbf{36.0}      &   \textbf{19.2 }     \\
    \bottomrule
    \end{tabular}
    \end{footnotesize}
    \caption{\textbf{ObjectNav validation results in Simulation (Gibson-1 and -2).} We compare fully supervised, partially self-supervised, and fully self-supervised methods with our approach. We perform competitively with SOTA fully supervised methods on Gibson-1 and significantly outperforms end-to-end self supervised methods on Gibson-2. Cells with N.A. are due to gaps in the literature; ZER and ZSON only report results on Gibson-2, while the other methods only do so on Gibson-1. $*$ stands for our own implementation of these methods, since the original code of \cite{poni} that reports results of SemExp and PONI on Gibson-1 was not fully available.}
    \label{tab:objectnav_results}
    \vspace{-2em}
\end{table}

\subsection{Error Modes}
\label{subsec:errormodes}
\begin{figure}[!t]
    \centering
    \includegraphics[width=0.5 \textwidth]{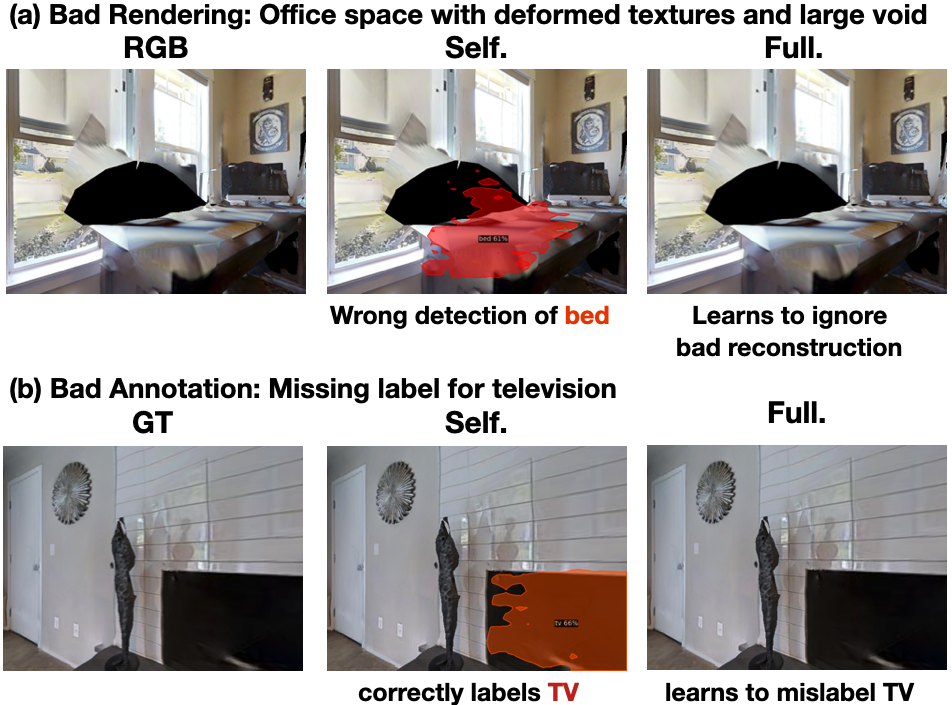}
    \vspace{-1.5em}
    \caption{\textbf{Data challenges: bad rendering and annotations.}}
    \label{fig:data_challenges}
    \vspace{-1em}
\end{figure}

Table~\ref{tab:nav_errormode} shows the breakdown of the error modes of \textit{FullSup} agent (denoted Full.) and \textit{SelfSup} agent (denoted Self.).
We find that simulation reconstruction and annotation can be very noisy from certain views and can produce RGB that are highly unlikely in the real world (Fig.~\ref{fig:data_challenges}), 
\begin{wrapfigure}[11]{r}{0.5\linewidth}
\vspace{-5pt}
\begin{center}
\begin{footnotesize}
\fontsize{7}{7}\selectfont
\setlength\tabcolsep{3pt} 
    \noindent \begin{tabular}{@{}@{\hspace{2pt}} l rr@{}}
         \toprule
          \textbf{Error Modes}& \textbf{Self.}  & \textbf{Full.}    \\
         \toprule
         \multicolumn{1}{@{}l}{\textbf{Misrecognition}} & 13.2\%  & 16.0\%   \\ 
          \midrule
         \multicolumn{1}{@{}l}{\textbf{\textit{Bad Rendering}$^*$}} & \textit{\textbf{8.0\%}}  & \textit{\textbf{1.2\%}} \\
         \midrule
         \multicolumn{3}{@{}l}{\textbf{The Rest}} \\
          Policy Error & 3.6\% & 2.4\%       \\
          Mapping Error & 6.0\% & 7.2\%  \\
          \textit{Mislabeled Goal}$^*$& 5.6\% & 5.2\%   \\
          Depth Render Error & 1.6\% &  0.8\%  \\
          $\Rightarrow$ Total  &  16.8\% & 15.6\%  \\
         \bottomrule
    \end{tabular}
\end{footnotesize}
\end{center}
\captionof{table}{\textbf{Error modes of Gibson-1 ObjectNav.} $^*$: Error modes \textit{unique to simulation}. }
\label{tab:nav_errormode}
\vspace{3em}
\end{wrapfigure}
and first enumerate the errors that are \textit{unique to simulation}, highlighted with $*$ in Table~\ref{tab:nav_errormode}. 
These errors are \textit{Bad Rendering} (Fig.~\ref{fig:data_challenges}(a); deformed rendering causes Self. to incorrectly detect goal objects and incorrectly terminate Nav task) and \textit{Mislabeled Goal} (Fig.~\ref{fig:data_challenges}(b); the goal objects are incorrectly labeled). We find that Self. is more susceptible to these errors, with especially \textit{Bad Rendering} happening $\sim$7 times more in Self. than Full. While this gap essentially explains the performance gap between \textit{SelfSup} agent and fully supervised methods (Table~\ref{tab:objectnav_results}), we note that these errors are caused by simulation artifacts. This agrees with our result in \S\ref{sec:real_insitu}, that adapting to these errors \textit{harm} real-world transfer.

Among errors not caused by reconstruction/ annotation challenges, the most common error mode for both agents is \textit{misrecognition} - goal objects not recognized even though an instance of it was seen or misrecognized (e.g. ``couch'' as ``bed''); surprisingly, Self. is more robust to this error. The other errors consist of policy error (bad exploration led to not finding a goal object), mapping/ planning error (agent goes out of the map and gets lost), and depth render error (agent stops outside the 1m radius due to depth error).

\begin{figure}[!t]
\centering
\footnotesize
\fontsize{8}{8}\selectfont
\setlength\tabcolsep{3pt} 
\begin{tabular}{@{}lc@{\hspace{5pt}}c@{\hspace{5pt}}cc@{\hspace{5pt}}c@{\hspace{5pt}}c@{\hspace{5pt}}c@{}}
\toprule
\multirow{3}{*}{\hspace{0.5em}\textbf{Metric}} & \multicolumn{3}{c}{\textbf{Test with \textit{Sim Noise}}} && \multicolumn{3}{c}{\textbf{Test w.o. \textit{Sim Noise}}} \\[1pt] 
   \cmidrule{2-4}  \cmidrule{6-8} 
   & O-t-S. & Self.  & Full. && O-t-S. & Self. & Full. \\
\midrule

\; Mean IOU & 38.3 & 44.1 & 52.7 && 44.2 & 53.1  & 53.1 \\   %
\; Freq. IOU
                & 92.2 & 93.0 & 94.0 &&  94.9 & 95.6  &  95.6 \\  %
\; Pixel Acc.
                & 95.1 & 95.9 & 96.5 && 96.8 & 97.4  & 97.4 \\    %

\bottomrule

\end{tabular}
\captionof{table}{\textbf{Semantic segmentation results on simulated data.} Results are on the 6 goal objects (IOUs of all 15 objects are in Table~\ref{tab:sim-seg-obj}); \textit{Self-Seg} and \textit{Full-Seg} are denoted Self. and Full.}
\label{tab:sim-seg}
\vspace{-2em}
\end{figure}

\subsection{Full. vs Self. Semantic Segmentation} 
\label{subsec:seg_comparison}
We further analyze the behavior of \textit{Self-Seg} (denoted Self.) and \textit{Full-Seg} (denoted Full.), to better understand the results of Table~\ref{tab:nav_errormode}. We present results on two test sets, which each consist of 500 images from 5 validation houses of Gibson. We realize that simply randomly teleporting agents and collecting images (Test with \textit{Sim Noise}) results in many images with bad rendering (that will not appear in the real world) and bad annotations (Fig.~\ref{fig:data_challenges}). Thus, we further evaluate on a test set of 500 images with obvious reconstruction/ annotation errors filtered by the authors (Test without \textit{Sim Noise}).

Table~\ref{tab:sim-seg} shows that \textit{Self-Seg} (Self.) yields performance on par with \textit{Full-Seg} (Full.) across all metrics, when evaluated without simulation noise. However, on Test with \textit{Sim Noise}, the performance of Self. sharply drops while that of Full. nearly stays the same. This leads to the conclusion that supervised training with reconstruction noise leads to learning simulation noise and artifacts; in Fig~\ref{fig:data_challenges}, we observe that Full. learns to both ignore bad renderings and adapt to the bad annotations of simulation. This explains the $\sim$7x gap in the \textit{Bad Rendering} error in Table~\ref{tab:nav_errormode} and the result that Full. harms real-world transfer, presented in \S\ref{sec:real_insitu}.

Table \ref{tab:sim-seg-obj} shows the class IOU of the three models across object categories. 
Interesting cases are bolded, but we draw the reader's attention in particular to cases where full supervision has hurt performance. 
This appears to happen with small objects (\eg  \textit{Cup} and \textit{Book}); we find these objects are often mislabeled. 
Because \textit{Self.} does not use labels, we are unaffected by this noise -- a good sign for more realistic scenarios. 
However, at present, small objects pose a major challenge for all models regardless of their supervision.

\begin{figure}[!t]
\centering
\fontsize{8}{8}\selectfont
\setlength\tabcolsep{4pt} 
\begin{tabular}{@{}l@{\hspace{5pt}}ccc@{}}
\toprule
\textbf{Category}
   &\textbf{O-t-S.} & \textbf{Self} & \textbf{Full} \\
\midrule
\;Chair
               & 41.9 &  40.3 & 42.6 \\    %
\;Couch
                & 57.7 &  60.7 & 62.4  \\    %
\;Potted Plant
                & 25.0  & 29.6 &  30.1 \\    %
\;Bed
                & 49.0  & 64.8 & 75.8  \\    %
\;Toilet
                 & \textbf{5.53 } & \textbf{ 37.1 }&  \textbf{48.7} \\    %
\;TV
                &  \textit{\textbf{32.7}} &  \textit{\textbf{41.5} }& \textit{\textbf{14.5}}  \\    %
\;Dining Table
                & 11.0  & 11.9 & 19.6  \\    %
\;Oven
                 & 17.6  & 35.1 & 54.4  \\    %
               
\bottomrule
\end{tabular}
~
\begin{tabular}{@{}l@{\hspace{5pt}}ccc@{}}
\toprule
\textbf{Category}
   &\textbf{O-t-S.} & \textbf{Self} & \textbf{Full} \\
\midrule
\;Sink
                & 17.6 & 25.7  &  34.8 \\    %
\;Refrigerator
                & 25.3  & 36.7 & 45.2 \\    %

\;Vase
                  & 0.88  & 1.79 & 4.93 \\  %
\;Bottle
                 & 0.00 & 0.22 &  0.00 \\    %
\;Clock
                & 0.00  & 0.00 &  0.00 \\    %
\;Cup
                  & \textit{\textbf{28.4}}  & \textit{\textbf{33.2}} & \textit{\textbf{0.00} } \\   %
\;Book
                &  \textit{\textbf{11.2}} & \textit{\textbf{5.33}} & \textit{\textbf{1.18}} \\    %
\;\textbf{Total Mean}
                 & 27.4 & 32.5 & 33.2 \\    %
\bottomrule
\end{tabular}
\captionof{table}{\textbf{Semantic segmentation by object category}. We provide a category level IOU breakdown of results in Table \ref{tab:sim-seg}. 
\textit{Self.} shows a significant boost over \textit{O-t-S.} for \textbf{Toilet}, Bed, and Oven; Full. can hurt performance sometimes 
(\eg \textit{\textbf{Book}}, \textit{\textbf{Cup}}, \textit{\textbf{TV}}). }
\label{tab:sim-seg-obj}
\vspace{-2em}
\end{figure}

\subsection{Full. vs Self. Nav Policy}
\label{subsec:polcomparison}
First, we investigate if the gap of \textit{policy error} in Table~\ref{tab:nav_errormode} is meaningful. In Table~\ref{tab:objectnav_ablations}, we compare ObjectNav performance on Gibson - 1 with a cross product of \textit{two segmentation models} and \textit{three ObjectNav policy models}; the former are \textit{Self-Seg} and \textit{Full-Seg}, and the latter are \textit{Self-Pol}, \textit{Full-Pol}, and SemExp \cite{ogn} trained with ground truth locations from labeled 3d meshes.
\setlength{\columnsep}{4pt}
\captionsetup{font=tiny}
\begin{wrapfigure}[9]{r}{0.5\linewidth}
   \centering
    \footnotesize
    \fontsize{8}{8}\selectfont
    \setlength\tabcolsep{3pt} 
    \begin{tabular}{@{}lccc@{}}
    
    \toprule
       & \multicolumn{3}{c}{\textbf{Segmentation Models}} \\
       \cmidrule{2-4}
    \textbf{Nav Policy}   & Self-Seg & Full-Seg \\
    \midrule
    
    \; Self-Pol & 60.0/ 31.2  & 64.4/ 33.2  \\
    \; Full-Pol\footnotemark\!\!\!
                    & 56.3/ 28.9 &  58.8/ 28.8 \\
    \; SemExp
                    & 59.0/ 32.3 & 63.4/ 32.2    \\   
    
    \bottomrule
    
    \end{tabular}
    \captionof{table}{
    \fontsize{8.3}{8.3}\selectfont  ObjectNav \textit{``Success Rate/ SPL''}  on Gibson-1 for \textit{segmentation models} with different \textit{navigation policy models}.}
    \label{tab:objectnav_ablations}
\end{wrapfigure}
Table~\ref{tab:objectnav_ablations} shows that \textit{Self-Pol} performs as well as \textit{Full-Pol} across segmentation models, implying that existing expensive methods that train the Nav policy with self-supervision, such as using ground truth locations of objects, collecting demonstrations, or using ImageNav (in which the goal location comes for free as mentioned in \cite{zson}) are not needed.

Furthermore, this result implies that policy training is fairly agnostic to noise; in Table~\ref{tab:nav_errormode}, we see that \textit{Self-Seg} introduces more noise to the self-labeled semantic map (from \textit{Bad Rendering}) than the mislabeling in the mesh (from \textit{Mislabeled Goal}). Still, \textit{Self-Pol}, while trained with more noisy data, performs on par with (even slightly better than) its counterparts trained with mesh annotations. We hypothesize that this is because the Nav Policy is trained to capture abstract knowledge on house layouts (as mentioned in \cite{ogn, poni}) and thus is fairly robust to both noises in Fig.~\ref{fig:data_challenges}.

\footnotetext{The lower performance of Full-Sup PONI may be because of misalignment between the pre-trained model provided by \cite{poni} and our reimplementation of the navigation components which we could not reproduce with the provided code. We also compare against Full-Sup SemExp.}

\section{Conclusion}
We present a self-supervised method to train all components of an ObjectNav agent that works in the real world. We find that our method can provide robust transfer to the real world, while supervised training with simulation annotations can actually harm transfer performances. Most importantly, we show that our method can work for \textit{in-situ} training to further improve ObjectNav performance, which encourages future research efforts on \textit{in-situ} training. Our aim is to further encourage the development of physical robot systems that adapt without requiring human intervention.

\bibliographystyle{unsrt}
\bibliography{main.bib}

\end{document}